\documentclass[runningheads]{llncs}
\usepackage[T1]{fontenc}
\usepackage{graphicx}
\usepackage{hyperref}
\usepackage{color}
\usepackage{subcaption}
\usepackage{amsmath}

\begin{document}
\title{A Multiscale Network with Supervised Contrastive Learning for Real-Time Facial Emotion Recognition}

%
%
\author{Rejoy Chakraborty\inst{1}\orcidID{0000-0001-9584-1838} \and
Archisman Adhikary\inst{2}\orcidID{0009-0005-1411-3465} \and
Chayan Halder\inst{3}\orcidID{0000-0002-6113-7284} \and
Payel Rakshit\inst{4}\orcidID{0000-0002-9006-2437} \and
Sanchita Ghosh\inst{5}\orcidID{0000-0002-1410-0024}
\and
Kaushik Roy\inst{5}\orcidID{0000-0002-3360-7576}}
\authorrunning{R. Chakraborty et al.}
%
\institute{Indian Statistical Institute, Kolkata, WB, India \\
\email{rejoychakraborty929@gmail.com}\\
\and
Department of Biological Sciences, Bose Institute, Kolkata, 700091, WB, India\\
\email{archismanadhikary@gmail.com}\\
\and
Ramakrishna Mission Vivekananda Centenary College, Kolkata, WB, India\\
\email{chayan.halderz@gmail.com}\\
\and
Maheshtala College, Kolkata, WB, India \\
\email{prmylife20@gmail.com}\\
\and
West Bengal State University, Barasat, WB, India \\
\email{\{psysanchita, kaushik.mrg\}@gmail.com}}
\maketitle              
\begin{abstract}
Real-time emotion recognition from facial expressions is a challenging task, particularly in video-based scenarios where multiple emotional states may occur over time. The difficulty increases further due to the fact that each emotional state is associated with facial expressions that vary significantly across individuals. The change of facial expressions portraying emotional state is not discrete, but rather continuous, which is very challenging to represent through computational aids. A system with the ability to detect variations in facial expressions can have a significant impact on determining the emotional state of an individual. Such a system can be very beneficial for psychologists during counseling by providing additional insights into the emotional state of a subject. In this paper, a deep learning-based system is presented to detect emotional changes in real-time video of a person by modeling the change in facial expressions. The current study is conducted on a standard dataset for training of the deep learning system and the system has provided very satisfactory outcomes in this respect.

\end{abstract}
\section{Introduction}
Emotion Recognition is one of the currently active research topics that deals with human emotions through an artificial recognition system. Emotion recognition technology can provide benefits to various fields like healthcare, education, and marketing by providing insights into individuals' emotional states. The facial expression~\cite{sariyanidi2014automatic}, vocal tone~\cite{anagnostopoulos2015features}, psychological signals~\cite{shu2018review}, etc., can be used to get insight into one's emotional state. Mehrabian et. al.~\cite{marechal2019survey} have analyzed that emotion recognition through facial expression can be one of the most impactful approaches.\par
Emotions and psychological processes are deeply connected ~\cite{koval2015emotional,majid2012role}. The relationship between emotional and psychological changes is bidirectional, which indicates that changes in emotional state can also influence psychological processes. For example, experiencing positive emotions, such as happiness or joy, can lead to an increase in motivation and engagement in activities, which can have a positive impact on psychological functioning, such as improved self-esteem or sense of purpose. To recognize and understand the emotional and psychological state of a person, we have experts in that domain who can understand the significance of the changes in emotions. Now, in day-to-day life, the ratio between psychological experts and people suffering from psychological issues is dreadful. As a consequence, very few people are getting the help of the experts. Here, an automated AI-based system to determine the emotional changes from facial expressions will be crucial for society. 

Facial expressions can often be ambiguous, with different emotions sometimes being conveyed by similar-looking expressions. For example, a smile naturally indicates happiness, but it can also refer to hiding negative emotions or showing politeness. Other than that, a facial expression can be heavily influenced by the context in which it is observed. For example, a smile in one situation may indicate happiness, while in another it may signal sarcasm or insincerity. Hence, emotions are not static and can change rapidly, often with subtle shifts in facial expression. Changes in emotion can occur over different time scales. Some changes may be relatively short-lived, and some may continue for a day or longer. Due to these factors, it is challenging to accurately recognize emotions in real-time, especially when dealing with video or live-streaming data.

\par Several researchers have proposed their models regarding emotion recognition through facial expression, which are mostly Convolutional Neural Network (CNN)-based architectures ~\cite{mollahosseini2016going,li2018occlusion,mohammadpour2017facial,yolcu2019facial}. To the best of our knowledge, none of the previous research works have focused on analyzing and monitoring emotional changes over an extended period, which is crucial for comprehensively understanding a person's psychological state. Simply recognizing emotions in a single image cannot provide conclusive information about someone's mental health. Thus, our research addresses this gap by providing a detailed analysis of emotional changes over time. The key contributions of the study are mentioned below: 

\begin{enumerate}
  \item Proposed MSFERNet, a multi-scale Facial Emotion recognition architecture, outperformed standard datasets FER13~\cite{goodfellow2013challenges} and CK+~\cite{lucey2010extended}.
  \item Developed RT-FER, a user-friendly application for real-time emotion monitoring.
  \item The GUI has incorporated a video player, which will provide the user with an opportunity to capture changes in emotion due to the effect of the video.
\end{enumerate}

\par The rest of the paper is organized into different sections. The recent study of literature is presented in Section \ref{Sec:related_work}. The dataset used for the proposed study is described in Section \ref{Sec: data}. An extensive description of the methodology is provided in Section \ref{Sec: method}. Section \ref{Sec:Results} provides the experimental results and analysis of the study. The paper is concluded in section \ref{Sec:conclusion}.

\section{Literature Review}
\label{Sec:related_work}
Numerous researchers have proposed their architectures to achieve high accuracy in recognizing emotions through facial expressions, trained using standard datasets like FER13~\cite{goodfellow2013challenges}, MMI~\cite{pantic2005web}, CK+~\cite{lucey2010extended}, RAFD-DB~\cite{li2017reliable} etc. In particular, many of them have also focused on the preprocessing stage to enhance the feature extraction process for facial expressions.\par
Mollahosseini et al.~\cite{mollahosseini2016going} proposed a CNN model for facial recognition using facial landmark extraction and applied data augmentation techniques for regularization. The architecture stacks two modules similar to Inception module~\cite{szegedy2015going} including pooling layers and it achieves high recognition rates of $94.7\%$, $77.9\%$, $55\%$, $76.7\%$, $47.7\%$, $93,2\%$, and $61.1\%$ on standard datasets MultiPie~\cite{gross2010multi}, MMI~\cite{pantic2005web}, DISFA~\cite{mavadati2013disfa}, FERA~\cite{valstar2011first}, SFEW~\cite{dhall2011static}, CK+~\cite{lucey2010extended}, FER13~\cite{goodfellow2013challenges} respectively. Li et al.~\cite{li2018occlusion} proposed a VGGNet~\cite{simonyan2014very} based network, followed by applying the ACNN technique with an attention mechanism while achieving an accuracy of $80.54\%$ and $54.84\%$ on RaFD~\cite{langner2010presentation} and AffectNet~\cite{mollahosseini2017affectnet} respectively. Mohammadpour et al.~\cite{mohammadpour2017facial} introduced a new approach for identifying the facial Action Units (AU) using shallow CNN architecture with only two convolutional layers followed by max-pooling layers and two dense layers identifying the number of AUs. Preprocessing has been applied before feeding to the network. The network architecture consisted of two convolution layers, each followed by max pooling, and two fully connected layers at the end to predict the number of activated AUs and achieved an accuracy of $97.01\%$ on CK+~\cite{lucey2010extended} dataset.\par

Jain et al.~\cite{jain2019extended} proposed a ConvoCNN architecture that integrates two residual blocks, each consisting of four convolution layers. The model was trained on CK+~\cite{lucey2010extended} and JAFFE~\cite{lyons1998japanese} databases, performing cropping and intensity normalization as preprocessing steps, and achieved $93.24\%$ and  $95.23\%$, recognition rates, respectively. Kim et al.~\cite{kim2017multi} proposed a spatio-temporal architecture that combines CNN and LSTM to address the variation of facial expressions, where CNN was first used to learn the spatial features of the facial expressions in all frames of the emotional state, and an LSTM was used to preserve the sequence of these spatial features. The proposed architecture shows a performance of $78.61\%$, and $60.98\%$ over the datasets MMI~\cite{pantic2005web} and CASME II~\cite{yan2014casme}, respectively. Punuri et al.~\cite{punuri2023efficient} proposed a model composition of EfficientNet~\cite{tan2019efficientnet} and XGBoost~\cite{ogunleye2019xgboost}, which is trained and evaluated using CK+~\cite{lucey2010extended}, FER13~\cite{goodfellow2013challenges},JAFFE~\cite{lyons1998japanese} and KDEF~\cite{jain2018hybrid} and received test accuracy of $94.41\%$, $61.54\%$, $97.67\%$ and $93.74\%$ respectively. \par
The study shows that there remains a significant gap in the current state of the work, where emotional change over a time frame has not been addressed. This could be useful in identifying an individual's psychological state. The proposed system will be the key to bridging the gap.

\section{Dataset}
\label{Sec: data}
\begin{figure}
  \centering
  \begin{subfigure}[b]{0.48\textwidth}
    \includegraphics[width=\textwidth]{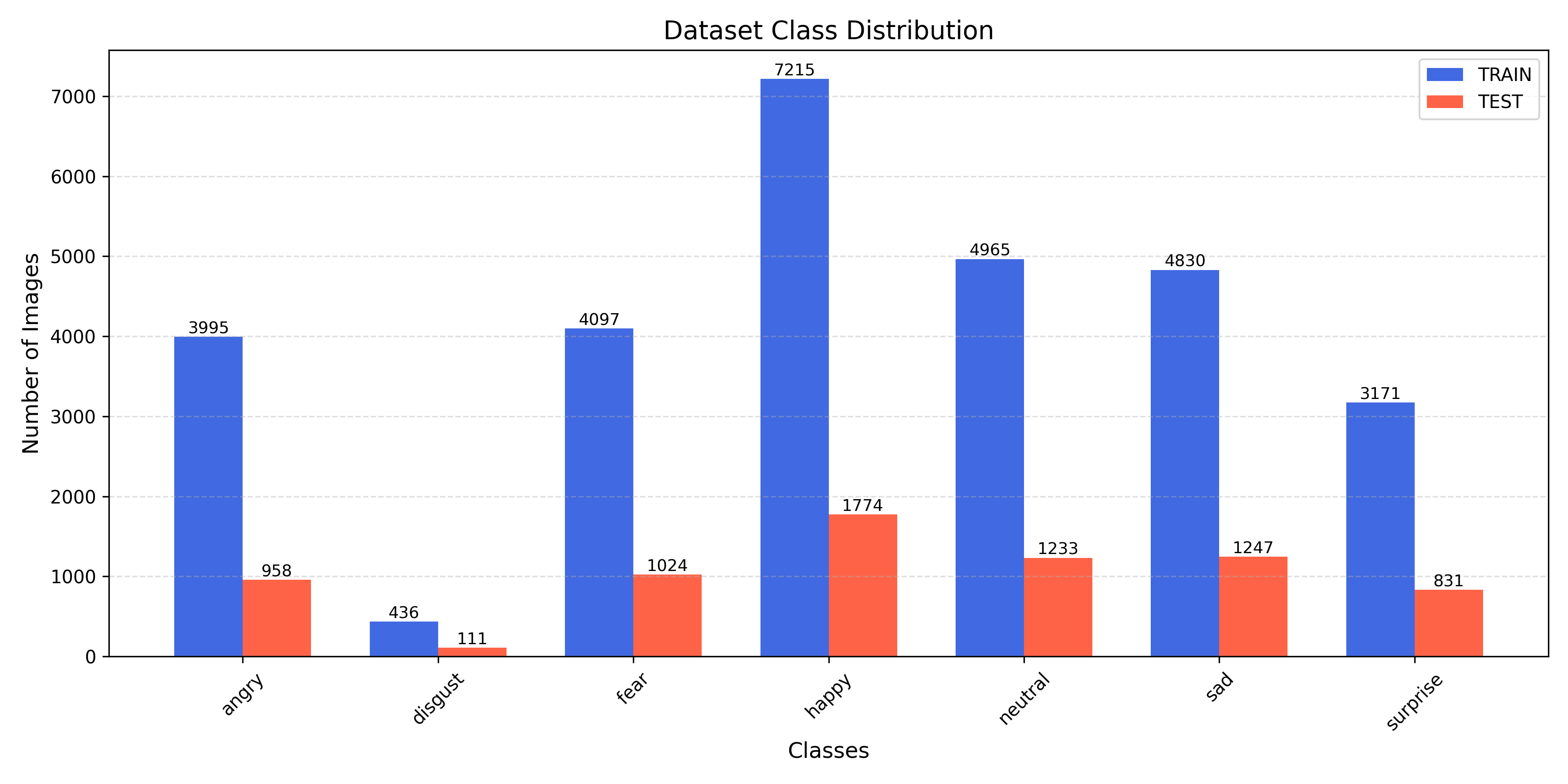}
    \caption{FER13-Original (7 class)}
    \label{subfig:orig_dist}
  \end{subfigure}
  \hfill
  \begin{subfigure}[b]{0.48\textwidth}
    \includegraphics[width=\textwidth]{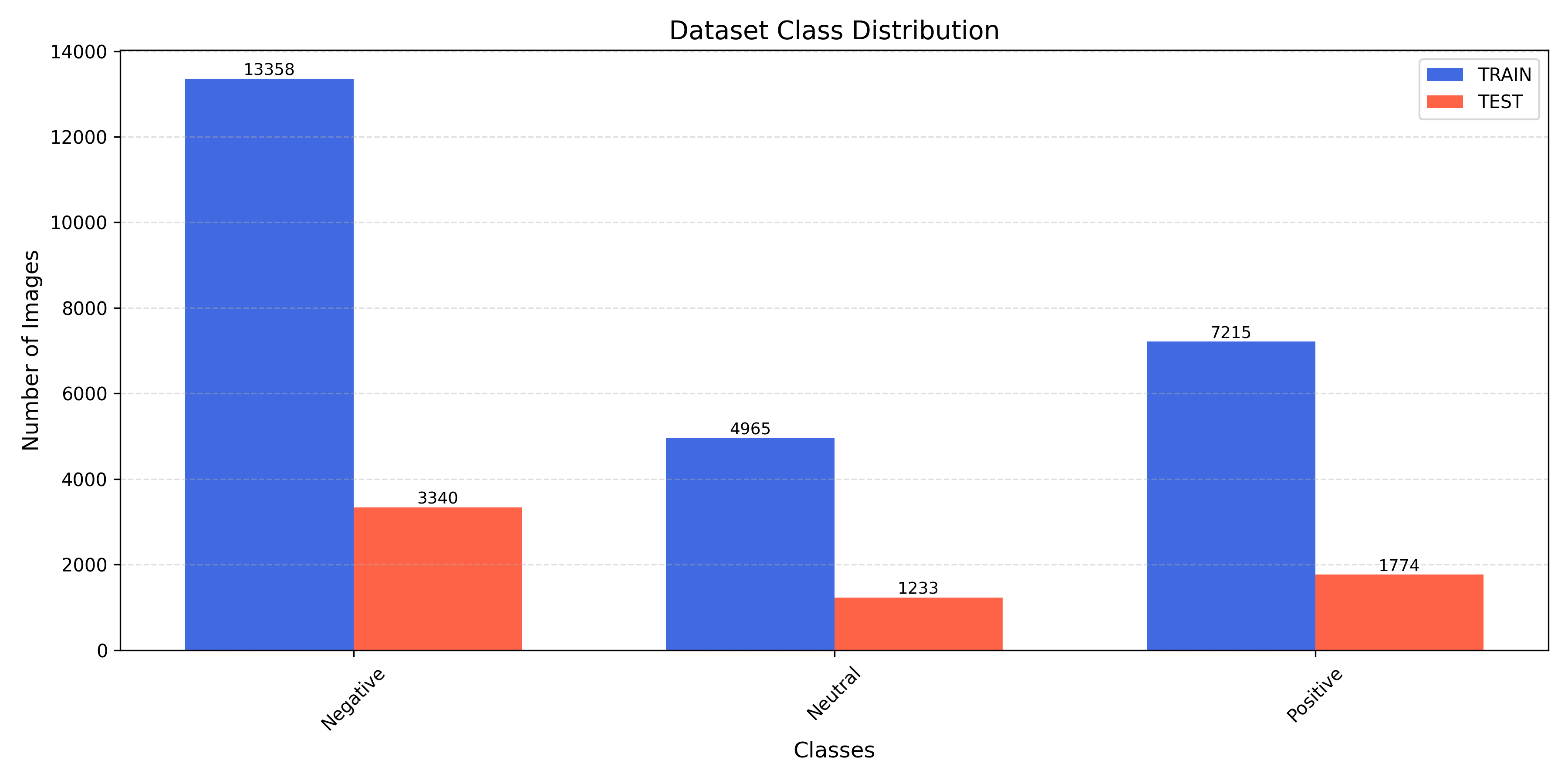}
    \caption{FER13-Modified (3 class)}
    \label{subfig:modi_dist}
  \end{subfigure}\\
  \vfill
  \begin{subfigure}[b]{0.48\textwidth}
    \includegraphics[width=\textwidth]{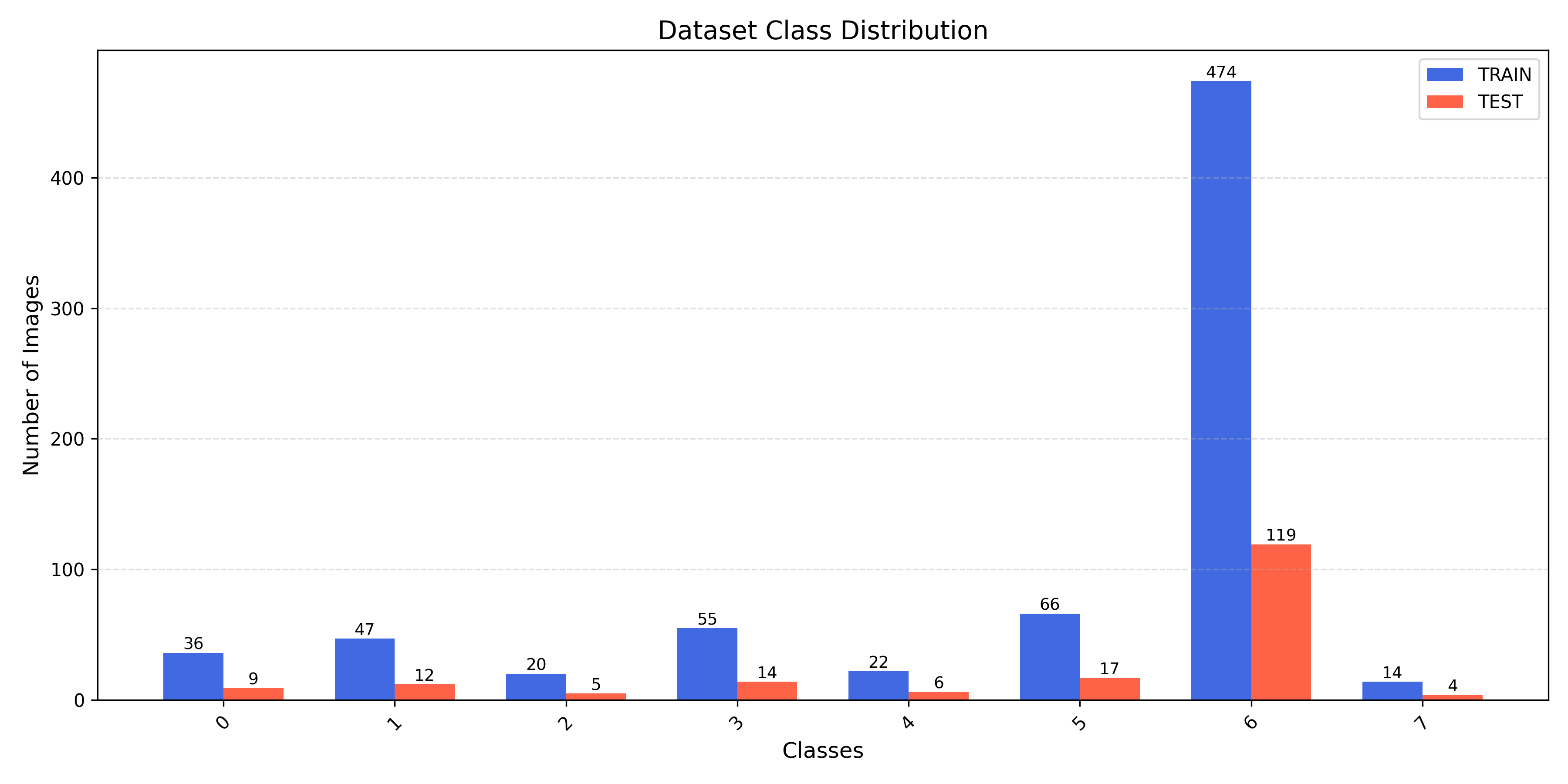}
    \caption{CK+}
    \label{subfig:ck_dist}
  \end{subfigure}
  \caption{Data distribution of the datasets}
  \label{fig:dist}
  
\end{figure}

Choosing a robust dataset is quite crucial for building an automated real-time system. FER13~\cite{goodfellow2013challenges}, and CK+~\cite{lucey2010extended} are two of the widely used standard datasets in the domain of emotion recognition. The FER13 and the CK+ datasets contain images of 7 and 8 classes, respectively. The data distributions have been shown in the figure \ref{subfig:orig_dist} and the figure \ref{subfig:ck_dist}, respectively.\par

In our current study, the main motivation is to provide a user-friendly system to psychologists that will aid the experts in easily identifying the mental state of their subjects. In this endeavor, the strongest emotions are the only effective ones. The total set of emotions is then categorized as `Positive', `Neutral', and `Negative' emotions. To the best of our knowledge, there is no such standard dataset that consists of data separated into these three emotional classes. To mitigate this, the seven emotions of FER13 are merged into 3 emotions, namely `Positive', `Negative', and `Neutral'. It has been curated by logically merging the classes `Angry', `Disgust', `Fear', and `Sad' into the class `Negative'. The `Positive' class contains the samples of the `Happy' class of the original, and the `Neutral' class is the same as it is. We purposely neglected the `Surprise' class as it is highly contextual and evokes mixed emotions. Moreover, basic preprocessing has been conducted to automatically eliminate unreadable or corrupted images of the dataset. Figure \ref{subfig:modi_dist} provides the distribution of the modified dataset.

\section{Methodology}
\label{Sec: method}
\begin{figure}[http]
    \centering
    \includegraphics[width=\textwidth]{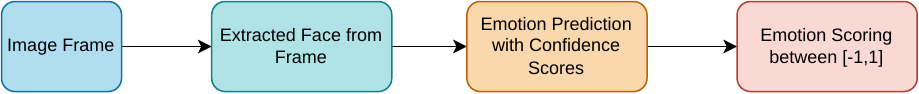}
    \caption{Emotion recognition of each frame}
    \label{fig:emotio_recog}
\end{figure}

The working principle of the proposed Emotion Monitoring system consists of two phases. In the first phase, it extracts the face from the frame, while the second phase performs classification to recognize the emotion. Figure \ref{fig:emotio_recog} depicts the workflow of the proposed system. To elaborate on both phases, this section is split into two major subsections. The first part elaborates on the proposed \textbf{MSFERNet}, used in the system for facial expression recognition, and the second part describes the interface of the emotion monitoring system \textbf{RT-FER}.

\subsection{MSFERNet: Multi-Scale Facial Expression Recognition Network}\label{subsec:2}

\begin{figure}[!htp]
    \centering
    \includegraphics[width=\textwidth]{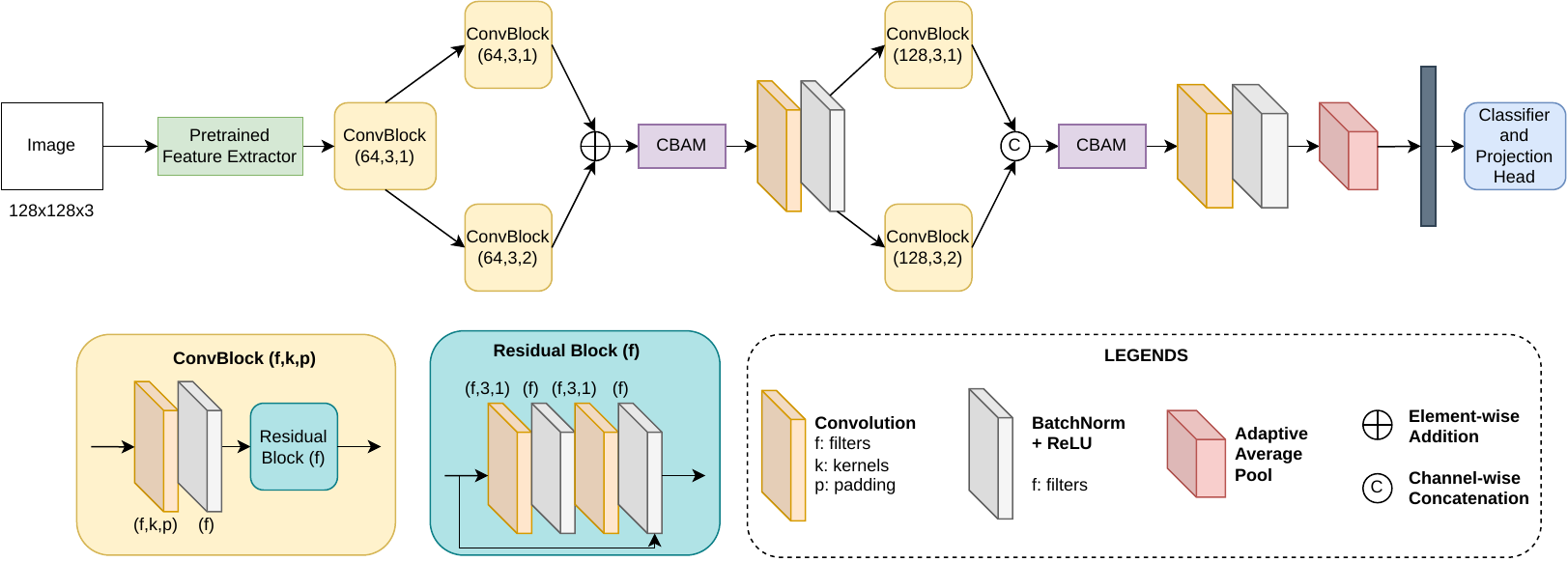}
    \caption{Proposed MSFERNet architecture design}
    \label{fig:CNN}
\end{figure}

Traditional sequential CNN often suffers from feature degradation and vanishing gradient problems as the network depth increases. To address these limitations, a multi-scale attention-based architecture, \textbf{M}ulti-\textbf{S}cale \textbf{F}acial \textbf{E}xpression \textbf{R}ecognition \textbf{Net}work (\textbf{MSFERNet}) has been proposed. The architecture incorporates transfer learning, residual mechanisms, attention mechanisms, and multi-scale feature extraction to improve discriminative facial representation learning while maintaining computational efficiency. The overall architecture of the proposed MSFERNet is illustrated in Figure~\ref{fig:CNN}. The proposed network takes an RGB facial image as input. Prior to training, several preprocessing and augmentation techniques, including width shifting, height shifting, zooming, shearing, horizontal flipping, and rescaling, were applied for data regularization and to improve model generalization. Initially, shallow feature representations are extracted using the first four stages of a pre-trained EfficientNet-B0 backbone~\cite{tan2020efficientnetrethinkingmodelscaling}. Instead of utilizing the complete EfficientNet architecture, only the early feature extraction layers are employed to capture low-level and mid-level semantic facial features while reducing computational complexity. The extracted feature maps are then refined through a convolutional refinement block consisting of a $3\times3$ convolution layer, batch normalization, ReLU activation, and a residual block. The residual block contains two consecutive $3\times3$ convolutional layers with skip connections to preserve identity mappings and reduce feature degradation during training. Residual learning also improves gradient flow and stabilizes deeper feature extraction. Following refinement, the network performs multi-scale feature extraction through parallel convolutional branches. In the first multi-scale stage, two parallel branches consisting of $3\times3$ and $5\times5$ convolutional kernels are utilized to capture both fine-grained local facial textures and broader spatial emotional patterns. The outputs of these branches are combined through element-wise addition and passed through a Convolutional Block Attention Module (CBAM)~\cite{woo2018cbam}. The CBAM module sequentially applies channel attention and spatial attention to emphasize informative facial regions and suppress irrelevant background features.

The refined features are then downsampled using a strided convolution layer and forwarded to a second multi-scale stage. Similar to the first stage, parallel $3\times3$ and $5\times5$ convolutional branches are employed. However, in this stage, the extracted feature maps are concatenated channel-wise to preserve complementary information from different receptive fields. Another CBAM attention module is subsequently applied to enhance high-level semantic facial representations. After attention-guided feature extraction, a convolutional downsampling layer is used followed by global average pooling to generate compact global feature descriptors. The pooled features are then passed through two fully connected layers with $128$ and $64$ hidden units, respectively. ReLU activation is applied after each fully connected layer, while a dropout layer with a dropout probability of $0.3$ is incorporated after the first fully connected layer to reduce overfitting. To improve representation learning, a projection head is additionally introduced for supervised contrastive learning. The projection head maps the extracted features into a normalized embedding space using two fully connected layers and feature normalization. Finally, the classification head predicts the facial emotion category through a linear layer followed by the Softmax activation function.

\subsection{RT-FER: Emotion Monitoring System}\label{subsec:3}
\begin{figure}
  \centering
  \begin{subfigure}[b]{0.8\textwidth}
    \includegraphics[width=\textwidth]{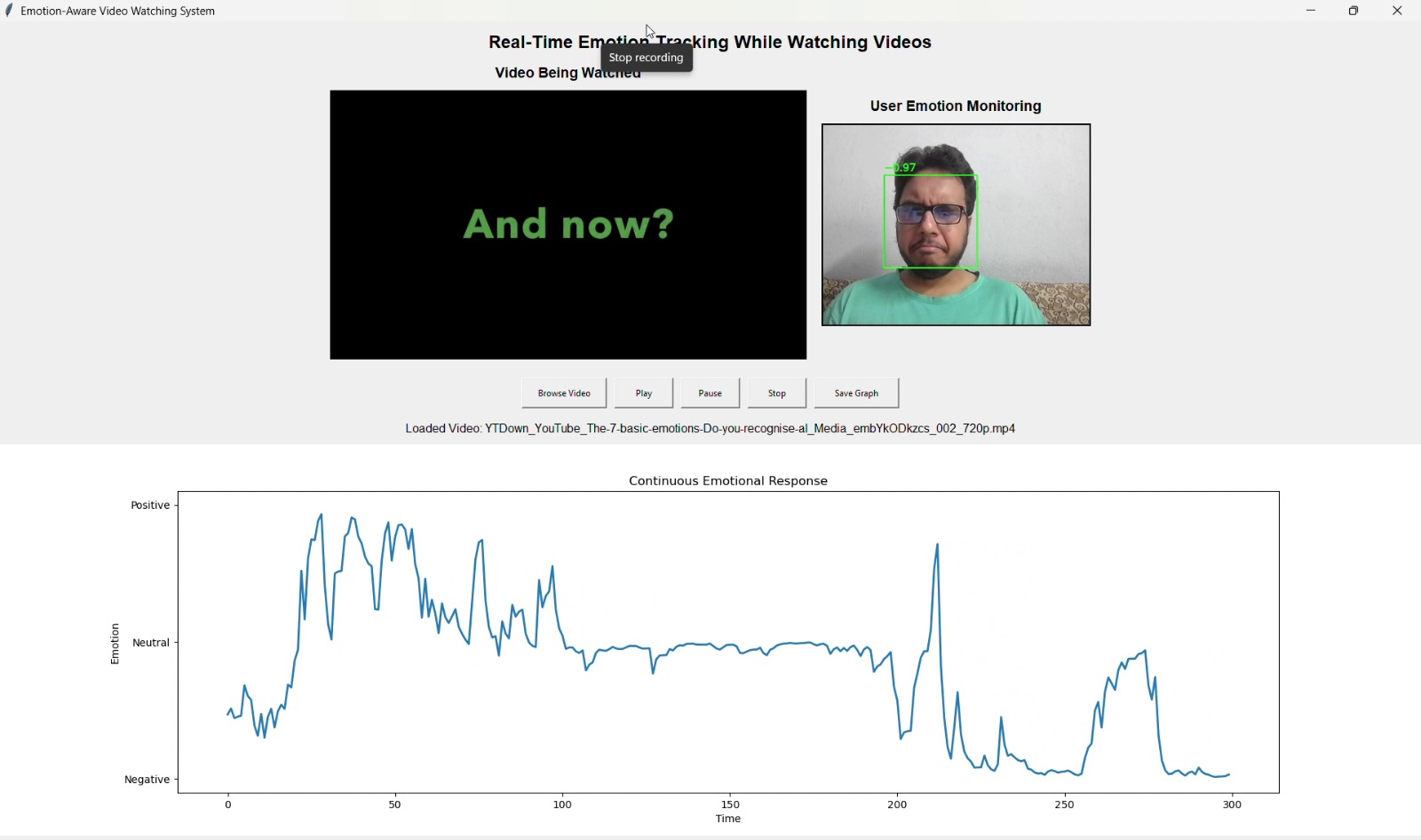}
    \caption{Sample Frame 1}
    \label{subfig:EMS_frame1}
  \end{subfigure} \\
  \begin{subfigure}[b]{0.8\textwidth}
    \includegraphics[width=\textwidth]{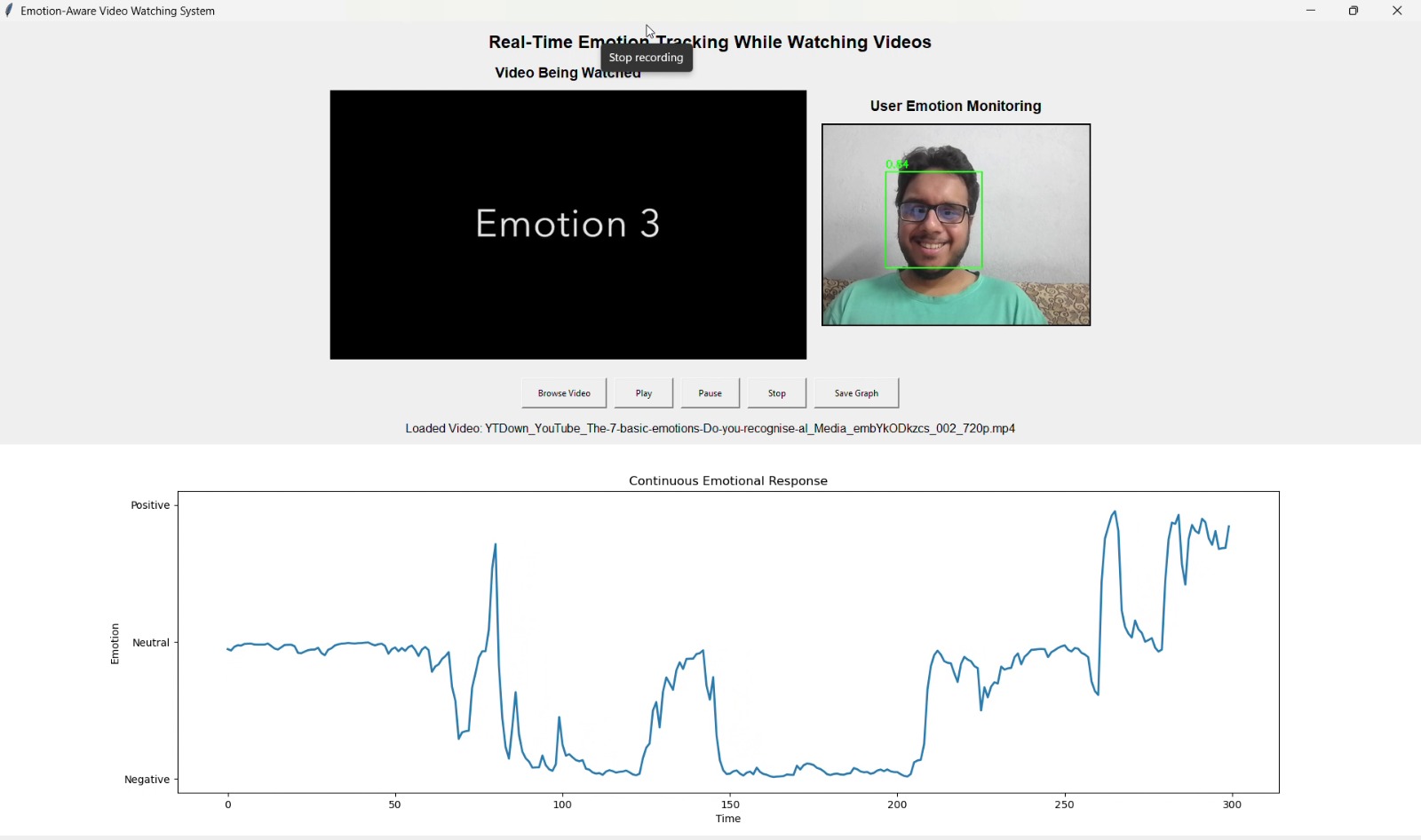}
    \caption{Sample Frame 2}
    \label{subfig:EMS_frame2}
  \end{subfigure}
  \caption{Sample view of proposed system RT-FER.}
  \label{fig:system}
\end{figure}

In order to track and monitor the emotional change of a person, an Emotion monitoring system, named \textbf{R}eal-\textbf{T}ime \textbf{F}acial \textbf{E}motion \textbf{R}ecognition \textbf{(RT-FER)} has been developed. A view of the system is visualized in Figure \ref{fig:system}. The system is provided for the research community at the URL \url{https://github.com/rejoycs/msfernet-emotion-monitor}. However, it can only be accessed upon request.\par

The system records live video from the user's perspective, treating it as a series of sequential frames. The image frame may contain irrelevant content other than the face. Hence, the face extraction operation has been performed on each frame. After the face extraction, each ROI image of the face is reshaped into $128 \times 128$ and passed to the proposed trained MSFERNet model to predict the emotion. The interface of the system is divided into two sections. One section contains the video parts and another one contains the analytical part. By browsing, a video can be selected which will be observed by the user, and that video will be displayed on the system itself. Other than that, the captured real-time video of the user will also be displayed on the system. While observing the video, the emotion will be predicted from the facial expression of the user. In the analytical part, a graph will track the emotion score over a certain period. In order to calculate the emotion score, we first assign weight values -1,0,1 to ``Negative'', ``Neutral'' and ``Positive''. Let the model return a class probability vector.  The emotion can be defined as:
\[
Score = (-1)*p_{\textbf{negative}} + 0*p_{\textbf{neutral}} + 1*p_{\textbf{positive}}
.\]
Hence essentially the score is $p_{\textbf{positive}}-p_{\textbf{negative}}$. Users can play/pause/stop the system anytime and can also save the generated plot. The labels mentioned in the graphs ``Positive'' and ``Negative'' indicate extremely or purely ``Positive'' emotion and purely ``Negative'' emotion.\par

\section{Results and Analysis}
\label{Sec:Results}
\subsection{Experimental Setup}\label{subsec:expsetup}
The model has been trained and evaluated using FER13~\cite{goodfellow2013challenges} with both original and modified versions, along with CK+~\cite{lucey2010extended}. All the images has been resized to $128 \times 128$. In all the cases, the dataset has been split into $80:10$ ratio for training and testing.

For classification, the model has been trained using Categorical Cross Entropy Loss $\mathcal{L}_{\text{cross}}$, defined as

\[
\mathcal{L}_{\text{cross}}
=
-\frac{1}{N}
\sum_{i=1}^{N}
\sum_{c=1}^{C}
y_{ic}\log(\hat{y}_{ic})
\]

where $N$ denotes the batch size, $C$ represents the number of emotion classes, $y_{ic}$ is the ground truth label, and $\hat{y}_{ic}$ is the predicted probability for class $c$.

To learn better representations of emotional features, the model has further trained in a contrastive framework using Supervised Contrastive Loss~\cite{khosla2020supervised} $\mathcal{L}_{\text{sup}}$, formulated as

\[
\mathcal{L}_{\text{sup}}
=
\sum_{i \in I}
\frac{-1}{|P(i)|}
\sum_{p \in P(i)}
\log
\frac{
\exp \left( \mathbf{z}_i \cdot \mathbf{z}_p / \tau \right)
}{
\sum_{a \in A(i)}
\exp \left( \mathbf{z}_i \cdot \mathbf{z}_a / \tau \right)
}
\]

where $\mathbf{z}_i$ and $\mathbf{z}_p$ denote the normalized feature embeddings of anchor and positive samples respectively, $\tau$ is the temperature parameter, $P(i)$ represents the set of positive samples belonging to the same class as sample $i$, and $A(i)$ denotes all samples excluding the anchor itself.

The total loss function is defined as

\[
\mathcal{L}
=
1.0 \times \mathcal{L}_{\text{cross}}
+
0.1 \times \mathcal{L}_{\text{sup}}
\]

The pipeline has been optimized using AdamW optimizer with a learning rate of 0.0002 and a weight decay of 0.0001, with a batch size of 128.
\par

\subsection{Experimental Results}
\begin{figure}[http]
    \centering
    \includegraphics[width=\textwidth]{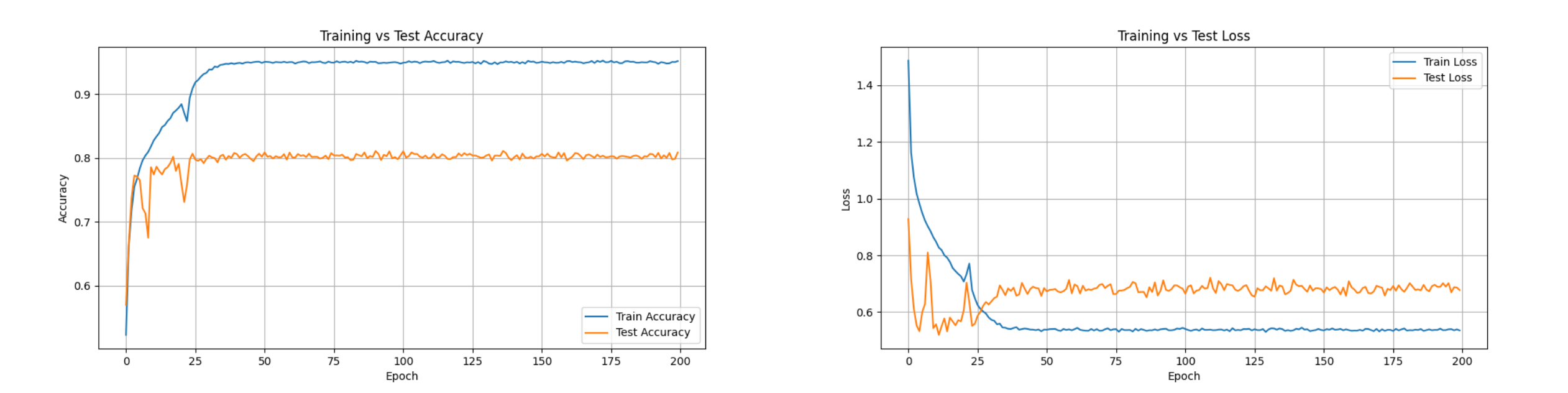}
    \caption{The Accuracy and Loss curve over the epochs of MSFERNet on the modified FER13 with 3 classes.}
    \label{fig:plot_3}
\end{figure}

The proposed MSFERNet model has been extensively evaluated on the FER13 dataset in both its original 7-class configuration and modified 3-class configuration, along with the CK+ dataset. The quantitative performance analysis, including classification accuracy and loss values across different datasets (test set), is summarized in Table~\ref{tab:eval_tab}. Further, the accuracy and loss curve have been shown in Figure \ref{fig:plot_3} of the modified FER13 dataset, which has been used in the RT-FER system.


\begin{table}[h]
\centering
\caption{Evaluation metrics across different datasets}
\label{tab:eval_tab}
\begin{tabular}{|c|c|c|}
\hline
Dataset & Test Accuracy (\%) & Test Loss \\
\hline
FER13 (7-class) & 66.73 & 0.98 \\
\hline
FER13 (3-class) & 81.08 & 0.52 \\
\hline
CK+ & 96.77 & 0.09 \\
\hline
\end{tabular}
\end{table}

In the current study, the results achieved for the 3-class experiment can not be directly compared, as similar attempts are not available in the literature. The comparative results of the MSFERNet with other state-of-the-art models are presented in Table \ref{tab:compare}. It can be seen that the proposed model not only outperforms the SOTA methods on both datasets, also resource efficient with only $2.37$M parameters. This helps for integrating the model in low resource device also.
\begin{table}[h]
\centering
\caption{Comparative study of the proposed model on FER2013 (7 class) and CK+ datasets}
\label{tab:compare}

\resizebox{\textwidth}{!}{
\begin{tabular}{|l|l|c|c|c|}
\hline
Ref & Model & Trainable Params. & FER2013 (7-class) & CK+ \\
\hline

Mollahosseini et al.~\cite{mollahosseini2016going} 
& AlexNet 
& $62.30$M 
& $61.10\%$ 
& $91.68\%$ \\
\hline

Mollahosseini et al.~\cite{mollahosseini2016going} 
& Proposed Model 
& $24.17$M 
& $65.08\%$ 
& $92.08\%$ \\
\hline 

Liu et al.~\cite{liu2016facial} 
& VGGNet 
& $84.00$M 
& $62.44\%$ 
& $91.39\%$ \\
\hline 

Punuri et al.~\cite{punuri2023efficient} 
& EfficientNet+XGBoost 
& $5.30$M  
& $61.54\%$ 
& $94.41\%$ \\
\hline

\textbf{Proposed} 
& \textbf{MSFERNet} 
& $\mathbf{2.37}$M 
& $\mathbf{66.73\%}$ 
& $\mathbf{96.77\%}$ \\
\hline

\end{tabular}
}
\end{table}

\begin{table}[h]
\centering
\caption{Impact of Supervised Contrastive (SupCon) Loss on classification accuracy (\%).}
\label{tab:eval_tab}
\begin{tabular}{|c|c|c|}
\hline
Dataset & w/ SupCon Loss & w/o SupCon Loss \\
\hline
FER13 (7-class) & 66.73 & 64.19 \\
\hline
FER13 (3-class) & 81.08 & 80.94 \\
\hline
CK+ & 96.77 & 95.56 \\
\hline
\end{tabular}
\end{table}

\subsection{Error Analysis}

\begin{figure}[http]
  \centering
  \begin{subfigure}[b]{0.3\textwidth}
    \includegraphics[width=\textwidth]{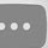}
    \caption{Training\_2167967.jpg}
    \label{subfig:error_1}
  \end{subfigure}
  \hfill
  \begin{subfigure}[b]{0.3\textwidth}
    \includegraphics[width=\textwidth]{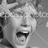}
    \caption{Training\_160942.jpg}
    \label{subfig:error_2}
  \end{subfigure}
  \caption{Some erroneous samples of FER13~\cite{goodfellow2013challenges}}
  \label{fig:error}
\end{figure}

The findings demonstrate satisfactory performance in handling the three-class scenario. Nonetheless, despite preprocessing efforts, certain erroneous images persisted, adversely impacting the model training. An error occurs when unwanted content comes into the image. For instance, in Figure \ref{subfig:error_1}, an image labeled "Training\_2167967.jpg" within the `Negative' training set doesn't even contain any face and instead features a logo commonly found in online media players due to technical glitches. Similarly, in Figure \ref{subfig:error_2}, "Training\_160942.jpg" displays a face but with a prominent watermark, introducing excessive noise. Moreover, the system will capture real-time images that vary in quality compared to the images on which the model was trained. Due to these scenarios, recognition becomes challenging.

\section{Conclusion}\label{Sec:conclusion}
This work presents an efficient and lightweight facial emotion recognition framework named \textbf{MSFERNet} for automated emotional state analysis from facial expressions. The proposed architecture integrates EfficientNet-based feature extraction, residual learning, multi-scale convolution, CBAM attention, and supervised contrastive learning to improve feature representation and classification performance. Experimental results demonstrate the effectiveness of the proposed model, achieving $66.73\%$ accuracy on the FER2013 seven-class dataset, $96.77\%$ accuracy on the CK+ dataset, and $81.08\%$ accuracy in the modified three-class FER2013. Despite its strong performance, the model contains only $2.37$ million trainable parameters, making it computationally efficient and suitable for real-time and mobile-based applications. The proposed system can assist in automated emotion monitoring and reduce the dependency on continuous manual observation. In future work, training with larger real-world datasets, incorporating stronger attention mechanisms, and extending the framework for real-time video-based emotion recognition may further improve system robustness and generalization performance.

\subsection*{Acknowledgements} We are grateful to the Department of Psychology of West Bengal State University for bringing the basic idea of the problem statement.

%
%
%

\bibliographystyle{splncs04}
\bibliography{references}

@article{sariyanidi2014automatic,
  title={Automatic analysis of facial affect: A survey of registration, representation, and recognition},
  author={Sariyanidi, Evangelos and Gunes, Hatice and Cavallaro, Andrea},
  journal={IEEE transactions on pattern analysis and machine intelligence},
  volume={37},
  number={6},
  pages={1113--1133},
  year={2014},
  publisher={IEEE}
}

@article{anagnostopoulos2015features,
  title={Features and classifiers for emotion recognition from speech: a survey from 2000 to 2011},
  author={Anagnostopoulos, Christos-Nikolaos and Iliou, Theodoros and Giannoukos, Ioannis},
  journal={Artificial Intelligence Review},
  volume={43},
  pages={155--177},
  year={2015},
  publisher={Springer}
}

@article{shu2018review,
  title={A review of emotion recognition using physiological signals},
  author={Shu, Lin and Xie, Jinyan and Yang, Mingyue and Li, Ziyi and Li, Zhenqi and Liao, Dan and Xu, Xiangmin and Yang, Xinyi},
  journal={Sensors},
  volume={18},
  number={7},
  pages={2074},
  year={2018},
  publisher={MDPI}
}

@article{marechal2019survey,
  title={Survey on AI-Based Multimodal Methods for Emotion Detection.},
  author={Marechal, Catherine and Mikolajewski, Dariusz and Tyburek, Krzysztof and Prokopowicz, Piotr and Bougueroua, Lamine and Ancourt, Corinne and Wegrzyn-Wolska, Katarzyna},
  journal={High-performance modelling and simulation for big data applications},
  volume={11400},
  pages={307--324},
  year={2019}
}

@inproceedings{liu2016facial,
  title={Facial expression recognition with CNN ensemble},
  author={Liu, Kuang and Zhang, Mingmin and Pan, Zhigeng},
  booktitle={2016 international conference on cyberworlds (CW)},
  pages={163--166},
  year={2016},
  organization={IEEE}
}

@article{koval2015emotional,
  title={Emotional inertia and external events: The roles of exposure, reactivity, and recovery.},
  author={Koval, Peter and Brose, Annette and Pe, Madeline L and Houben, Marlies and Erbas, Yasemin and Champagne, Dominique and Kuppens, Peter},
  journal={Emotion},
  volume={15},
  number={5},
  pages={625},
  year={2015},
  publisher={American Psychological Association}
}

@article{majid2012role,
  title={The role of language in a science of emotion},
  author={Majid, Asifa},
  journal={Emotion Review},
  volume={4},
  number={4},
  pages={380--381},
  year={2012},
  publisher={Sage Publications Sage UK: London, England}
}

@inproceedings{pantic2005web,
  title={Web-based database for facial expression analysis},
  author={Pantic, Maja and Valstar, Michel and Rademaker, Ron and Maat, Ludo},
  booktitle={2005 IEEE international conference on multimedia and Expo},
  pages={5--pp},
  year={2005},
  organization={IEEE}
}

@inproceedings{lucey2010extended,
  title={The extended cohn-kanade dataset (ck+): A complete dataset for action unit and emotion-specified expression},
  author={Lucey, Patrick and Cohn, Jeffrey F and Kanade, Takeo and Saragih, Jason and Ambadar, Zara and Matthews, Iain},
  booktitle={2010 ieee computer society conference on computer vision and pattern recognition-workshops},
  pages={94--101},
  year={2010},
  organization={IEEE}
}

@inproceedings{li2017reliable,
  title={Reliable crowdsourcing and deep locality-preserving learning for expression recognition in the wild},
  author={Li, Shan and Deng, Weihong and Du, JunPing},
  booktitle={Proceedings of the IEEE conference on computer vision and pattern recognition},
  pages={2852--2861},
  year={2017}
}

@article{langner2010presentation,
  title={Presentation and validation of the Radboud Faces Database},
  author={Langner, Oliver and Dotsch, Ron and Bijlstra, Gijsbert and Wigboldus, Daniel HJ and Hawk, Skyler T and Van Knippenberg, AD},
  journal={Cognition and emotion},
  volume={24},
  number={8},
  pages={1377--1388},
  year={2010},
  publisher={Taylor \& Francis}
}

@article{gross2010multi,
  title={Multi-pie},
  author={Gross, Ralph and Matthews, Iain and Cohn, Jeffrey and Kanade, Takeo and Baker, Simon},
  journal={Image and vision computing},
  volume={28},
  number={5},
  pages={807--813},
  year={2010},
  publisher={Elsevier}
}

@article{lyons1998japanese,
  title={The Japanese Female Facial Expression (JAFFE) Dataset},
  author={Lyons, Michael and Kamachi, Miyuki and Gyoba, Jiro},
  journal={The Images Are Provided at No Cost for Non-Commercial Scientific Research Only. If You Agree to the Conditions Listed Below, You May Request Access to Download},
  year={1998}
}

@article{yan2014casme,
  title={CASME II: An improved spontaneous micro-expression database and the baseline evaluation},
  author={Yan, Wen-Jing and Li, Xiaobai and Wang, Su-Jing and Zhao, Guoying and Liu, Yong-Jin and Chen, Yu-Hsin and Fu, Xiaolan},
  journal={PloS one},
  volume={9},
  number={1},
  pages={e86041},
  year={2014},
  publisher={Public Library of Science San Francisco, USA}
}

@inproceedings{dhall2011static,
  title={Static facial expression analysis in tough conditions: Data, evaluation protocol and benchmark},
  author={Dhall, Abhinav and Goecke, Roland and Lucey, Simon and Gedeon, Tom},
  booktitle={2011 IEEE international conference on computer vision workshops (ICCV workshops)},
  pages={2106--2112},
  year={2011},
  organization={IEEE}
}

@article{mollahosseini2017affectnet,
  title={Affectnet: A database for facial expression, valence, and arousal computing in the wild},
  author={Mollahosseini, Ali and Hasani, Behzad and Mahoor, Mohammad H},
  journal={IEEE Transactions on Affective Computing},
  volume={10},
  number={1},
  pages={18--31},
  year={2017},
  publisher={IEEE}
}

@inproceedings{mollahosseini2016going,
  title={Going deeper in facial expression recognition using deep neural networks},
  author={Mollahosseini, Ali and Chan, David and Mahoor, Mohammad H},
  booktitle={2016 IEEE Winter conference on applications of computer vision (WACV)},
  pages={1--10},
  year={2016},
  organization={IEEE}
}

@inproceedings{mohammadpour2017facial,
  title={Facial emotion recognition using deep convolutional networks},
  author={Mohammadpour, Mostafa and Khaliliardali, Hossein and Hashemi, Seyyed Mohammad R and AlyanNezhadi, Mohammad M},
  booktitle={2017 IEEE 4th international conference on knowledge-based engineering and innovation (KBEI)},
  pages={0017--0021},
  year={2017},
  organization={IEEE}
}

@article{li2018occlusion,
  title={Occlusion aware facial expression recognition using CNN with attention mechanism},
  author={Li, Yong and Zeng, Jiabei and Shan, Shiguang and Chen, Xilin},
  journal={IEEE Transactions on Image Processing},
  volume={28},
  number={5},
  pages={2439--2450},
  year={2018},
  publisher={IEEE}
}

@article{yolcu2019facial,
  title={Facial expression recognition for monitoring neurological disorders based on convolutional neural network},
  author={Yolcu, Gozde and Oztel, Ismail and Kazan, Serap and Oz, Cemil and Palaniappan, Kannappan and Lever, Teresa E and Bunyak, Filiz},
  journal={Multimedia Tools and Applications},
  volume={78},
  pages={31581--31603},
  year={2019},
  publisher={Springer}
}

@article{jain2019extended,
  title={Extended deep neural network for facial emotion recognition},
  author={Jain, Deepak Kumar and Shamsolmoali, Pourya and Sehdev, Paramjit},
  journal={Pattern Recognition Letters},
  volume={120},
  pages={69--74},
  year={2019},
  publisher={Elsevier}
}

@article{kim2017multi,
  title={Multi-objective based spatio-temporal feature representation learning robust to expression intensity variations for facial expression recognition},
  author={Kim, Dae Hoe and Baddar, Wissam J and Jang, Jinhyeok and Ro, Yong Man},
  journal={IEEE Transactions on Affective Computing},
  volume={10},
  number={2},
  pages={223--236},
  year={2017},
  publisher={IEEE}
}

@article{mavadati2013disfa,
  title={Disfa: A spontaneous facial action intensity database},
  author={Mavadati, S Mohammad and Mahoor, Mohammad H and Bartlett, Kevin and Trinh, Philip and Cohn, Jeffrey F},
  journal={IEEE Transactions on Affective Computing},
  volume={4},
  number={2},
  pages={151--160},
  year={2013},
  publisher={IEEE}
}

@inproceedings{valstar2011first,
  title={The first facial expression recognition and analysis challenge},
  author={Valstar, Michel F and Jiang, Bihan and Mehu, Marc and Pantic, Maja and Scherer, Klaus},
  booktitle={2011 IEEE International Conference on Automatic Face \& Gesture Recognition (FG)},
  pages={921--926},
  year={2011},
  organization={IEEE}
}

@article{punuri2023efficient,
  title={Efficient net-XGBoost: an implementation for facial emotion recognition using transfer learning},
  author={Punuri, Sudheer Babu and Kuanar, Sanjay Kumar and Kolhar, Manjur and Mishra, Tusar Kanti and Alameen, Abdalla and Mohapatra, Hitesh and Mishra, Soumya Ranjan},
  journal={Mathematics},
  volume={11},
  number={3},
  pages={776},
  year={2023},
  publisher={MDPI}
}

@article{ogunleye2019xgboost,
  title={XGBoost model for chronic kidney disease diagnosis},
  author={Ogunleye, Adeola and Wang, Qing-Guo},
  journal={IEEE/ACM transactions on computational biology and bioinformatics},
  volume={17},
  number={6},
  pages={2131--2140},
  year={2019},
  publisher={IEEE}
}

@inproceedings{tan2019efficientnet,
  title={Efficientnet: Rethinking model scaling for convolutional neural networks},
  author={Tan, Mingxing and Le, Quoc},
  booktitle={International conference on machine learning},
  pages={6105--6114},
  year={2019},
  organization={PMLR}
}

@article{jain2018hybrid,
  title={Hybrid deep neural networks for face emotion recognition},
  author={Jain, Neha and Kumar, Shishir and Kumar, Amit and Shamsolmoali, Pourya and Zareapoor, Masoumeh},
  journal={Pattern Recognition Letters},
  volume={115},
  pages={101--106},
  year={2018},
  publisher={Elsevier}
}

@article{simonyan2014very,
  title={Very deep convolutional networks for large-scale image recognition},
  author={Simonyan, Karen and Zisserman, Andrew},
  journal={arXiv preprint arXiv:1409.1556},
  year={2014}
}

@inproceedings{szegedy2015going,
  title={Going deeper with convolutions},
  author={Szegedy, Christian and Liu, Wei and Jia, Yangqing and Sermanet, Pierre and Reed, Scott and Anguelov, Dragomir and Erhan, Dumitru and Vanhoucke, Vincent and Rabinovich, Andrew},
  booktitle={Proceedings of the IEEE conference on computer vision and pattern recognition},
  pages={1--9},
  year={2015}
}

@inproceedings{goodfellow2013challenges,
  title={Challenges in representation learning: A report on three machine learning contests},
  author={Goodfellow, Ian J and Erhan, Dumitru and Carrier, Pierre Luc and Courville, Aaron and Mirza, Mehdi and Hamner, Ben and Cukierski, Will and Tang, Yichuan and Thaler, David and Lee, Dong-Hyun and others},
  booktitle={International conference on neural information processing},
  pages={117--124},
  year={2013},
  organization={Springer}
}

@article{khosla2020supervised,
  title={Supervised contrastive learning},
  author={Khosla, Prannay and Teterwak, Piotr and Wang, Chen and Sarna, Aaron and Tian, Yonglong and Isola, Phillip and Maschinot, Aaron and Liu, Ce and Krishnan, Dilip},
  journal={Advances in neural information processing systems},
  volume={33},
  pages={18661--18673},
  year={2020}
}

@misc{tan2020efficientnetrethinkingmodelscaling,
      title={EfficientNet: Rethinking Model Scaling for Convolutional Neural Networks}, 
      author={Mingxing Tan and Quoc V. Le},
      year={2020},
      eprint={1905.11946},
      archivePrefix={arXiv},
      primaryClass={cs.LG},
      url={https://arxiv.org/abs/1905.11946}, 
}

@inproceedings{woo2018cbam,
  title={Cbam: Convolutional block attention module},
  author={Woo, Sanghyun and Park, Jongchan and Lee, Joon-Young and Kweon, In So},
  booktitle={Proceedings of the European conference on computer vision (ECCV)},
  pages={3--19},
  year={2018}
}
\end{document}